\documentclass[10pt,twocolumn,letterpaper]{article}

\usepackage{cvpr}
\usepackage{times}
\usepackage{epsfig}
\usepackage{graphicx}
\usepackage{amsmath}
\usepackage{amssymb}
\usepackage{float}
\usepackage{tabulary}
\usepackage{multirow}
\usepackage{booktabs}  
\usepackage{caption}
\usepackage{subfigure}
\usepackage{fancyhdr,graphicx}
\usepackage{diagbox}
\usepackage[lined,ruled,linesnumbered]{algorithm2e}

\usepackage[pagebackref=true,breaklinks=true,letterpaper=true,colorlinks,bookmarks=false]{hyperref}

 \cvprfinalcopy 

\def\cvprPaperID{5997} 

\DeclareMathOperator*{\argmin}{arg\,min}

\ifcvprfinal \fi
\begin{document}

\title{Learning Meta Face Recognition in Unseen Domains}

\author{Jianzhu Guo$^{1,2}$ \quad Xiangyu Zhu$^{1,2}$ \quad Chenxu Zhao$^{3}$ \quad Dong Cao$^{1,2}$ \quad Zhen Lei$^{1,2}$\thanks{Corresponding author} \quad Stan Z. Li$^{4}$\\
\small  $^{1}$ CBSR \& NLPR, Institute of Automation, Chinese Academy of Sciences, Beijing, China\\
\small  $^{2}$School of Artificial Intelligence, University of Chinese Academy of Sciences, Beijing 100049, China\\
\small	$^{3}$Mininglamp Academy of Sciences, Mininglamp Technology\\
\small	$^{4}$School of Engineering, Westlake University, Hangzhou, China\\
{\tt\small \{jianzhu.guo, xiangyu.zhu, dong.cao, zlei, szli\}@nlpr.ia.ac.cn, zhaochenxu@mininglamp.com} 
}

\maketitle
\thispagestyle{empty}

\begin{abstract}
    Face recognition systems are usually faced with unseen domains in real-world applications and show unsatisfactory performance due to their poor generalization. For example, a well-trained model on webface data cannot deal with the ID vs. Spot task in surveillance scenario.
    In this paper, we aim to learn a generalized model that can directly handle new unseen domains without any model updating.
    To this end, we propose a novel face recognition method via meta-learning named Meta Face Recognition (MFR).
    MFR synthesizes the source/target domain shift with a meta-optimization objective, which requires the model to learn effective representations not only on synthesized source domains but also on synthesized target domains.
    Specifically, we build domain-shift batches through a domain-level sampling strategy and get back-propagated gradients/meta-gradients on synthesized source/target domains by optimizing multi-domain distributions.
    The gradients and meta-gradients are further combined to update the model to improve generalization.
    Besides, we propose two benchmarks for generalized face recognition evaluation.
    Experiments on our benchmarks validate the generalization of our method compared to several baselines and other state-of-the-arts. The proposed benchmarks will be available at \url{https://github.com/cleardusk/MFR}.

\end{abstract}
\section{Introduction}

Face recognition is a long-standing topic in the research community. Recent works~\cite{taigman2014deepface,sun2014deep,schroff2015facenet,liu2017sphereface,deng2018arcface,wang2018support,wang2019co,wang2019mis} have pushed the performance to a very high level on several common benchmarks, e.g. LFW~\cite{huang2008labeled}, YTF~\cite{wolf2011face} and MegaFace~\cite{kemelmacher2016megaface}.
\begin{figure}[!htb]
    \centering
    \includegraphics[width=0.45\textwidth]{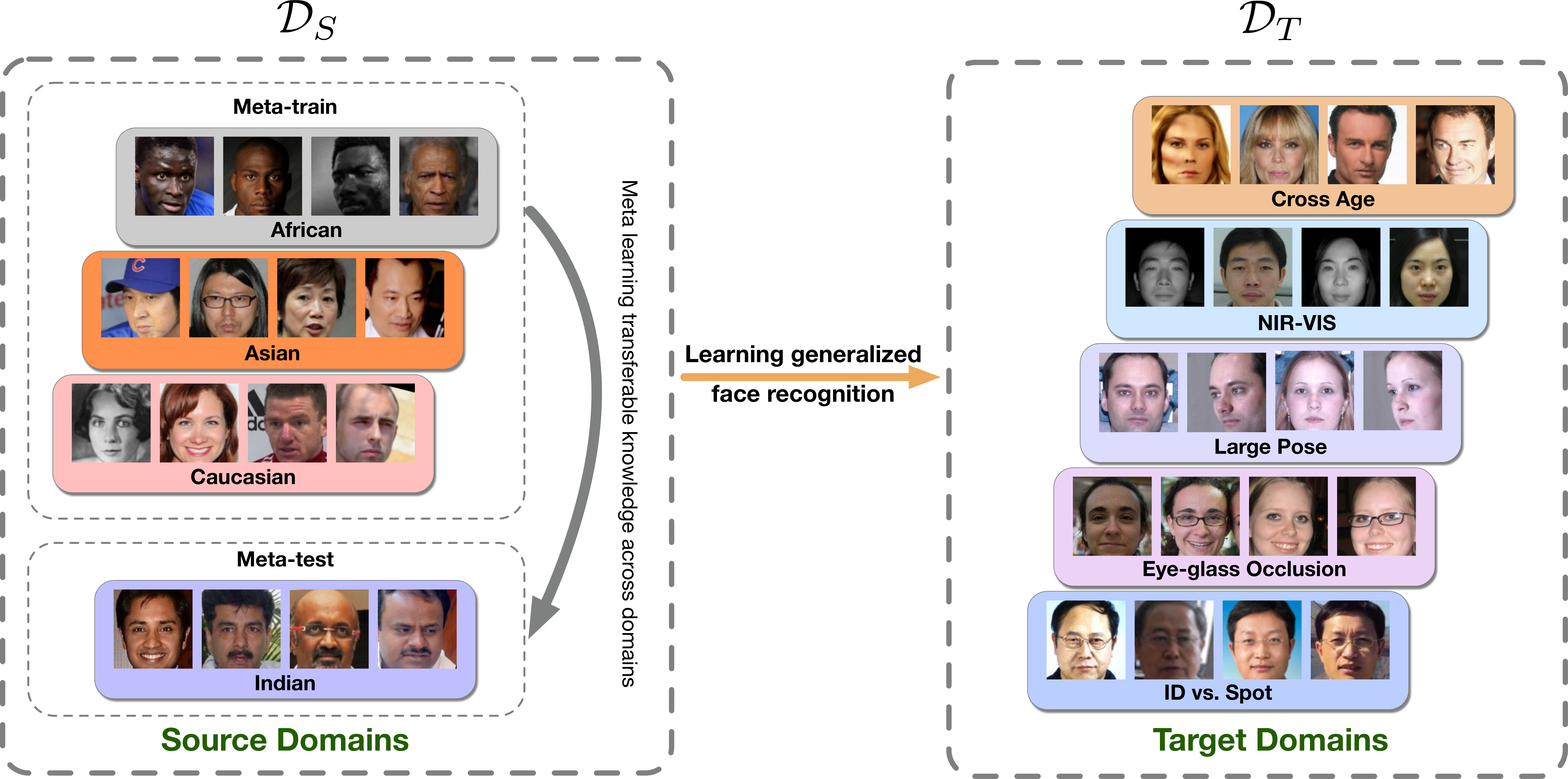}
    \caption{An illustration of our MFR for generalized face recognition problem. The left column contains four source domains of different races, the right includes five target domains: cross-age (CACD-VS), NIR-VIS face matching (CASIA NIR-VIS 2.0), large pose (Multi-PIE), eyeglass occlusion (MeGlass) and ID vs. Spot (Public-IvS), which are unseen in training. By meta-learning on the simulated meta-train/meta-test shifts in source domains, our model learns the transferable knowledge across domains to generalize well on target unseen domains.}
    \vspace{-1.5em}
    \label{fig_dg}
\end{figure}
These methods are based on the assumption that the training sets like CASIA-Webface~\cite{yi2014learning}, MS-Celeb~\cite{guo2016ms} and testing sets have similar distribution.
However, in real-world applications of face recognition, the model trained on source domains $\mathcal{D}_S$ is usually deployed in another domain $\mathcal{D}_T$ with a different distribution.
There are two kinds of scenarios: (i) the target domain $\mathcal{D}_T$ is known and the data is accessible. (ii) the target domain is unseen. Approaches to the first scenario are categorized into domain adaptation for face recognition, where the common setting is that the source domain $\mathcal{D}_S$ contains a labelled face domain and the target domain $\mathcal{D}_T$ is with or without labels. Domain adaption methods try to adapt the knowledge learned from $\mathcal{D}_S$ to $\mathcal{D}_T$ so that the model generalizes well on $\mathcal{D}_T$.
The second scenario can be regarded as domain generalization for face recognition, and we call it \textit{Generalized Face Recognition}, which is more common as the trained model is usually deployed in unknown scenarios and faced with unseen data. As illustrated in Fig.~\ref{fig_dg}, the deployed model should be able to generalize to unseen domains without any updating or fine-tuning.

Compared with domain adaptation, generalized face recognition is less studied and more challenging, since it makes no assumptions about target domains. To the best of our knowledge, there are no relative studies on generalized face recognition problem.
A related task is domain generalization on visual recognition, it assumes that the source and target domains share the same label space, and has a small set, e.g., 7 categories~\cite{li2018learning}.
However, generalized face recognition is an open-set problem and has a much larger scale of categories, making existing methods inapplicable.

In this paper, we aim to learn a model for generalized face recognition problem. Once trained on a set of source domains, the model can be directly deployed on an unseen domain without any model updating.
Inspired by~\cite{li2018learning,finn2017model}, we propose a novel face recognition framework via meta-learning named Meta Face Recognition (MFR).
MFR simulates the source/target domain shift with a meta-optimization objective, which optimizes the model to learn effective face representations not only on synthesized source domains but also on synthesized target domains.
Specifically, a domain-level sampling strategy is adopted to simulate the domain shift such that source domains are divided into meta-train/meta-test domains.
To optimize multi-domain distributions, we propose three components: 1) the hard-pair attention loss optimizes the local distribution with hard pairs, 2) soft-classification loss considers the global relationship within a batch and 3) domain alignment loss learns to reduce meta-train domains discrepancy by aligning domain centers. These three losses are combined to learn domain-invariant and discriminative face representations.
The gradients from meta-train domains and meta-gradients from meta-test domains are finally aggregated by meta-optimization, and are then used to update the network to improve model generalization.
Compared with traditional meta-learning methods, our MFR does not need model updating for target domains and can directly handle unseen domains.

Our main contributions include:
(i) For	the	first time,	we highlight the generalized face recognition problem, which requires a trained model to generalize well on unseen domains without any updating.
(ii) We propose a novel Meta Face Recognition (MFR) framework to solve generalized face recognition, which meta-learns transferable knowledge across domains to improve model generalization.
(iii) Two generalized face recognition benchmarks are designed for evaluation. Extensive experiments on the proposed benchmarks validate the efficacy of our method.
\section{Related work}

\textbf{Domain Generalization.} Domain generalization can be traced back to~\cite{khosla2012undoing,muandet2013domain}. DICA~\cite{muandet2013domain} adopts the kernel-based optimization to learn domain-invariant features. CCSA~\cite{motiian2017unified} can handle both domains adaptation and domain generalization problems by aligning a source domain distribution to a target domain distribution.
MLDG~\cite{li2018learning} firstly applies the meta-learning method MAML~\cite{finn2017model} for domain generalization.
Compared with domain adaptation, domain generalization is a less investigated problem. Besides, the above domain generalization works mainly focus on the closed-set category-level recognition problems, where the source and target domains share the same label space. In contrast, our generalized face recognition problem is much more challenging because the target classes are disjoint from the source ones. It means that generalized face recognition is an open-set problem rather than the closed-set problem like MLDG~\cite{li2018learning}, and we must handle the domain gap and the disjoint label space simultaneously.
One related work is DIMN~\cite{song2019generalizable}, but it differs from ours in both task and method. 

\begin{figure*}[!htb]
    \centering
    \includegraphics[width=0.99\textwidth]{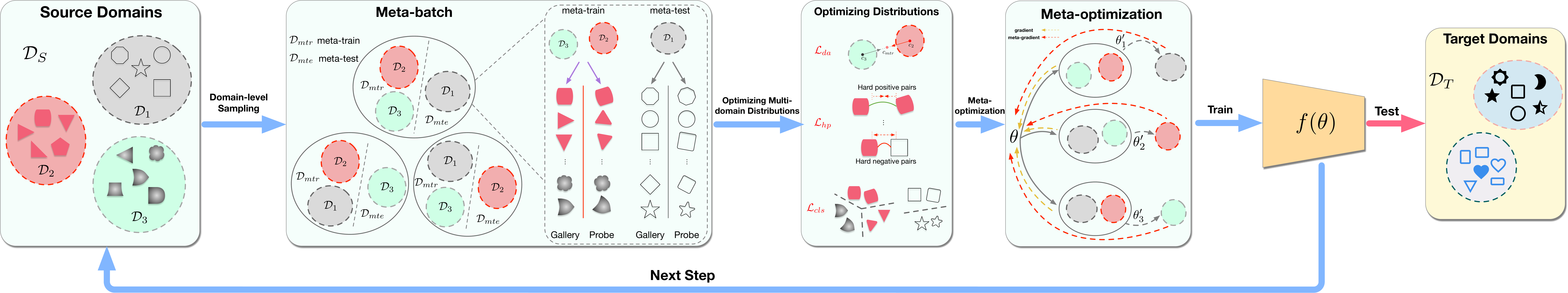}
    \vspace{-0.5em}
    \caption{Overview of our proposed MFR. Three source domains are presented in this figure for a demonstration. Each symbol represents a face image for convenience. MFR consists of three parts: domain-level sampling for simulating domain shifts, multi-domain distributions optimization to learn face representations and meta-optimization procedure to improve model generalization. Once trained on source domains, the model can be directly deployed on target unseen domains.}
    \vspace{-1em}
    \label{fig_sampling}
\end{figure*}

\textbf{Meta Learning.} Recent meta-learning studies concentrate on: (i) learning a good weight initialization for fast adaptation on a new task, such as the foundational work MAML~\cite{finn2017model} and its variants Reptile~\cite{nichol2018first}, meta-transfer learning~\cite{sun2019meta}, iMAML~\cite{rajeswaran2019meta} and so on. (ii) learning an embedding space with a well-designed classifier that can directly classify samples on a new task without fast adaptation~\cite{vinyals2016matching,snell2017prototypical,sung2018learning}. (iii) learning to predict the classification parameters~\cite{qiao2018few,gidaris2018dynamic} after pre-training a good feature extractor on the whole training set.
These works focus on few-shot learning, where the common setting is that the target task has very few data points (1/5/20 shots per class). In contrast, generalized face recognition should handle thousands of classes, making it more challenging and generally applicable.
Our approach is most related to MAML~\cite{finn2017model} that tries to learn a transferable weight initialization. However, MAML requires fast adaptation on a target task, while our MFR does not require any model updating as target domains are unseen.

\section{Methology}
This section describes the proposed MFR to address generalized face recognition problem. MFR consists of three parts: (i) the domain-level sampling strategy. (ii) three losses for optimizing multi-domain distributions to learn domain-invariant and discriminative face representations. (iii) the meta-optimization procedure to improve model generalization, shown in Fig.~\ref{fig_metaopt}. The overview is shown in Fig.~\ref{fig_sampling} and Algorithm~\ref{alg_mml}.

\subsection{Overview}
In the training stage, we have access to $N$ source domains $\mathcal{D}_S = \{ \mathcal{D}^S_1, \cdots, \mathcal{D}^S_N \mid N > 1\}$, and each domain $\mathcal{D}^S_i = \{(x^i_j, y^i_j) \}$ has its own label set. In the testing phase, the trained model is evaluated on one or several unseen target domains, $\mathcal{D}_T = \{ \mathcal{D}^T_1, \cdots, \mathcal{D}^T_M \mid M \geq 1 \}$, without any model updating.
Besides, the label sets $\mathcal{Y}_T$ of the target domains are disjoint from the label sets $\mathcal{Y}_S$ of source domains, thus making ours problem open-set.
During training, we define a single model represented by a parametrized function $f(\theta)$ with parameters $\theta$. Our proposed MFR aims to train $\theta$ on source domains $\mathcal{D}_S$, such that it can generalize well on target unseen domains $\mathcal{D}_T$, as illustrated in Fig.~\ref{fig_dg}.

\subsection{Domain-level Sampling}
To achieve domain generalization, we split source domains into meta-train and meta-test domains during each training iteration. Specifically, we split $N$ source domains $\mathcal{D}_S$ into $N-1$ domains $\mathcal{D}_{mtr}$ for meta-train and $1$ target domain $\mathcal{D}_{mte}$ for meta-test, simulating the domain shift problem existed when deployed in real-world scenarios. In this way, the model is encouraged to learn transferable knowledge about how to generalize well on the unseen domains with different distributions.
We further build a meta-batch consisting of several batches as follows:
(i) we iterate on $N$ source domains; (ii) in the $i$-th iteration, $\mathcal{D}_i^S$ is selected as the meta-test domain $\mathcal{D}_{mte}$; (iii) the rest ones as meta-train domains $\mathcal{D}_{mtr}$;
(iv) we randomly choose $B$ identities in meta-train domains and $B$ identities in meta-test domain, and two face images are selected for each identity, in which one as the gallery the other one as probe.
Therefore, a meta-batch of $N$ batches is built.
Our model is then updated by the accumulated gradients of each meta-batch. The details are illustrated in Algorithm~\ref{alg_mml}.
Different from MAML~\cite{finn2017model}, our sampling is domain-level for open-set face recognition.
MLDG~\cite{li2018learning} also performs a similar sampling, but their domains are randomly divided in each training iteration and no meta-batch is built.

\subsection{Optimizing Multi-domain Distributions}
To aggregate back-propagated gradients within each batch, we optimize multi-domain distributions such that the same identities are mapped into nearby representation and different identities are mapped apart from each other.
Traditional metric losses like contrastive~\cite{hadsell2006dimensionality,sun2014deep_iv} and triplet~\cite{schroff2015facenet} take randomly sampled pairs or triplets to build the training batches. These batches consist of lots of easy pairs or triplets, leading to the slow convergence of training.
To alleviate it, we propose to optimize and learn domain-invariant and discriminative representations with three components. The hard-pair attention loss optimizes the local distribution with hard pairs, the soft-classification loss considers the global distribution within a batch and the domain alignment loss learns to align domain centers.

\textbf{Hard-pair Attention Loss.}
Hard-pair attention loss focuses on optimizing hard positive and negative pairs.
A batch of $B$ identities are sampled and each identity contains a gallery face and a probe face. We denote the input as $\mathcal{X}$, the gallery and probe embeddings are extracted: $F_g = f(\mathcal{X}_{g} ;\theta) \in R^{B \times C}$, $F_p = f(\mathcal{X}_{p} ;\theta) \in R^{B \times C}$, where $C$ is the dimension length.
After $l_2$ normalization on $F_g$ and $F_p$, we can efficiently construct a similarity matrix by computing $M = F_g F_p^T \in R^{B \times B}$. Then we use a positive threshold $\tau_p$ and negative threshold $\tau_n$ to filter out the hard positive pairs and negative pairs: $\mathcal{P} = \{i | M_{i,i} < \tau_p  \}$ and $\mathcal{N} = \{ (i, j) | M_{i,j} > \tau_p, i \ne j  \}$. This operation just needs $O(B^2 \log(B))$ complexity and it can be formulated as:
\begin{equation}
    \label{eq_loss_cont_hard}
    \begin{split}
        \mathcal{L}_{hp} = \frac{1}{2|\mathcal{P}|} \sum_{i \in \mathcal{P}} \| F_{g_i} - F_{p_i}  \|_2^2 {-} \frac{1}{2|\mathcal{N}|} \sum_{(i, j) \in \mathcal{N}} \| F_{g_i} {-} F_{p_j}  \|_2^2,
    \end{split}
\end{equation}
where $\mathcal{P}$ is indices of hard positive pairs determined by $\tau_p$, $\mathcal{N}$ is indices of hard negative pairs determined by $\tau_n$.

\textbf{Soft-classification Loss.}
Hard-pair attention loss only concentrates on hard pairs and tends to converge to a local optimum. To alleviate it, we introduce a specific soft-classification loss to perform classification within a batch. The loss is formulated as:
\begin{equation}
    \label{eq_loss_cls}
    \begin{split}
        \mathcal{L}_{cls} {=} \frac{1}{2B} & \sum_{i=1}^{B} \Big( \text{CE} (y_i, s \cdot F_{g_i} W^T ) + \text{CE} (y_i, s \cdot F_{p_i} W^T ) \Big),
    \end{split}
\end{equation}
where $y_i = i$ indicates the i-th identity, $F_{g_i} W^T$ or $F_{p_i} W^T$ is the logit of i-th identity and $s$ is a fixed scaling factor. $W$ is initialized as $(F_g + F_p) / 2 \in R^{B \times C}$ and each row of $W$ is  $l_2$ normalized.

\begin{figure}[!htb]
    \centering
    \includegraphics[width=0.4\textwidth]{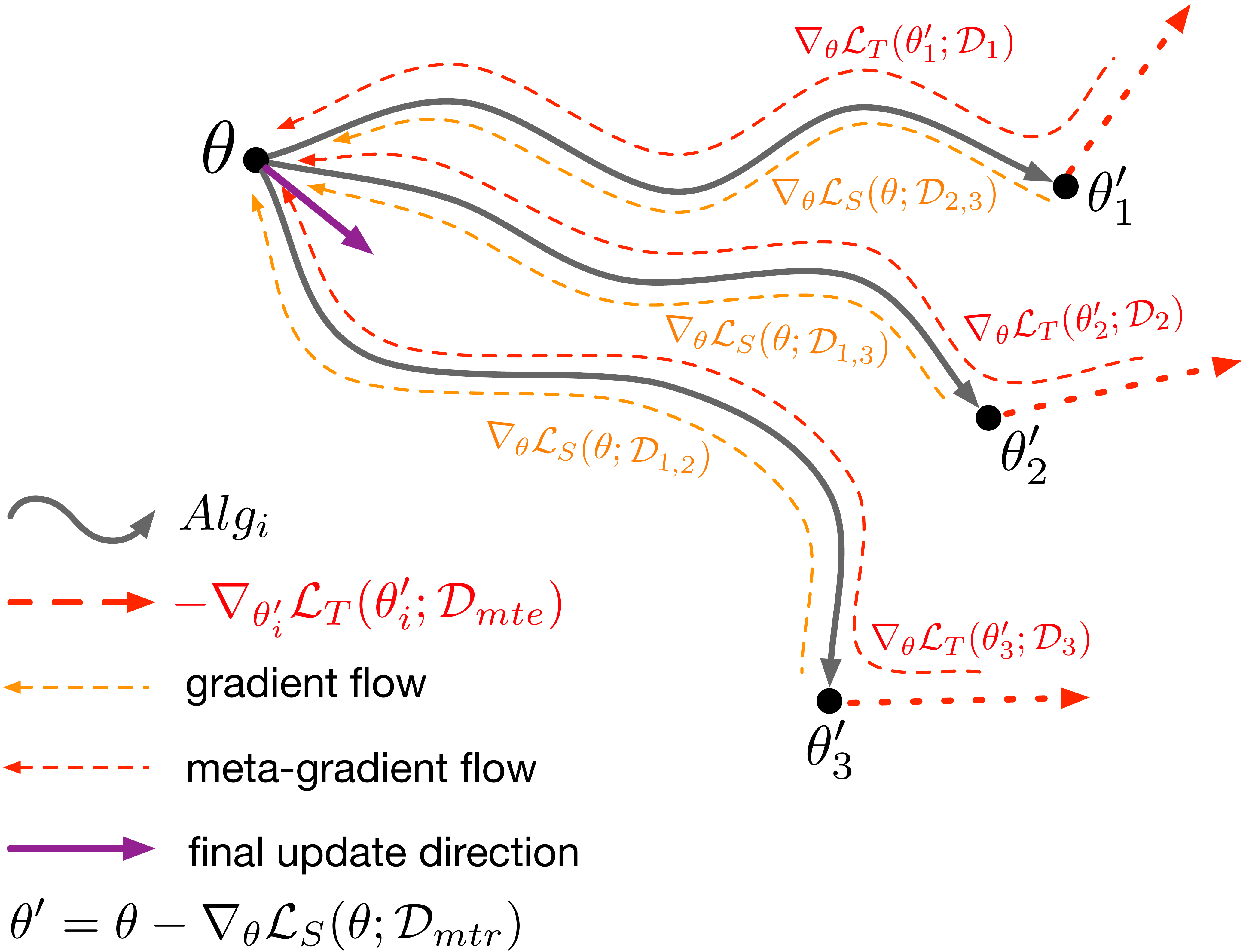}
    \vspace{-0.3em}
    \caption{Overview of the meta-optimization procedure in a meta-batch. Given three source domains $\mathcal{D}_{1,2,3}$, a meta batch contains three meta-train/meta-test divisions: $\mathcal{D}_{2,3}$ / $\mathcal{D}_{1}$, $\mathcal{D}_{1,3}$ / $\mathcal{D}_{2}$, $\mathcal{D}_{1,2}$ / $\mathcal{D}_{3}$. Each division contributes a gradient from meta-train and a meta-gradient from meta-test. The model is finally updated towards a direction that performs well on both meta-train and meta-test domains by accumulating all the gradients and meta-gradients.}
    \vspace{-0.5em}
    \label{fig_metaopt}
\end{figure}

\textbf{Domain Alignment Loss.}
We find negative pairs across meta-train domains tend to be easier than within domains. By adding a domain alignment regularization to make the embeddings domain-invariant, we can reduce domain gap of different meta-train domains. Besides, negative pairs across meta-train domains become harder, which is beneficial to learn more discriminative representations.
To perform domain alignment, we make the mean embeddings of multiple meta-train domains close to each other.
Specifically, we first calculate the embedding center of all mean embeddings of meta-train domains, then optimize the discrepancies between all mean embeddings and this embedding center.
The domain alignment loss is only applied on meta-train domains, as meta-test has only one domain. The loss is formulated as:
\begin{equation}
    \begin{split}
        c_j &= \frac{1}{B} \sum_{i=1}^{B} \big((F^{\mathcal{D}_j}_{g_i} + F^{\mathcal{D}_j}_{p_i}) / 2\big), \\
        c_{mtr} &= \frac{1}{n} \sum_{j=1}^{n} c_j, \\
        \mathcal{L}_{da} &= \frac{1}{n} \sum_{j=1}^{n} \| s \cdot (c_j - c_{mtr}) \|_2^2,
    \end{split}
\end{equation}
where $F_{g_i}$, $F_{p_i}$ are normalized embeddings, $c_j$ is the mean embedding within a batch sampled from domain $\mathcal{D}_j$, $c_{mtr}$ is the embedding center of all mean embeddings of meta-train domains, $n$ is the number of meta-train domains and $s$ is the scaling factor.
In meta-optimization, we will adaptively utilize the back-propagated signals from these three losses to improve the model generalization.

\subsection{Meta-optimization}
This section describes how the model is optimized to improve model generalization. The whole meta-optimization procedure is summarized in Algorithm~\ref{alg_mml} and illustrated in Fig.~\ref{fig_metaopt}.

\textbf{Meta-train.} Based on domain-level sampling, during each batch within a meta-batch, we sample $N-1$ source domains $\mathcal{D}_{mtr}$ and $B$ pairs of images $\mathcal{X}_S$ from $\mathcal{D}_{mtr}$. Then we conduct the proposed losses in each batch as follows:
\begin{equation}
\label{eq_metatrain}
	\mathcal{L}_S = \mathcal{L}_{hp}(\mathcal{X}_S; \theta) + \mathcal{L}_{cls} (\mathcal{X}_S; \theta) + \mathcal{L}_{da} (\mathcal{X}_S; \theta),
\end{equation}
where $\theta$ represents the model parameters. The model is next updated by gradient $\nabla_{\theta}$ as: $\theta' = \theta - \alpha \nabla_{\theta} \mathcal{L}_S (\theta)$. This update step is similar to the conventional metric learning.

\textbf{Meta-test.} In each batch, the model is also tested on the meta-test domain $\mathcal{D}_{mte}$. This testing procedure simulates the evaluating on an unseen domain with a different distribution, so as to make the the model to learn to generalize across domains. We also sample $B$ pairs of images $\mathcal{X}_T$ from the meta-test domain $\mathcal{D}_{mte}$. Then the loss is conducted on the updated parameters $\theta'$ as below:
\begin{equation}
	\label{eq_metatest}
	\mathcal{L}_T = \mathcal{L}_{hp} (\mathcal{X}_T; \theta') + \mathcal{L}_{cls} (\mathcal{X}_T; \theta').
\end{equation}

\textbf{Summary.}
To optimize the meta-train and meta-test simultaneously, the final MFR objective is:
\begin{equation}
	\label{eq_finalobj}
	\argmin_\theta  \gamma \mathcal{L}_S (\theta) +  (1-\gamma) \mathcal{L}_T (\theta - \alpha  \mathcal{L}'_S(\theta)),
\end{equation}
where $\alpha$ is the meta-train step-size and $\gamma$ balances meta-train and meta-test. This objective can be understood as: \textit{optimize the model parameters, such that after updating on the meta-train domains, the model also performs well on the meta-test domain}.
From another perspective, the second term of Eqn.~\ref{eq_finalobj} serves as an extra regularization to update the model with high order gradients, and we call it meta-gradients.
For example, given three source domains $\mathcal{D}_S = \{ \mathcal{D}_1, \mathcal{D}_2, \mathcal{D}_3\}$, a meta-batch consists of three meta-train/meta-test divisions: $\mathcal{D}_2, \mathcal{D}_3 / \mathcal{D}_1$, $\mathcal{D}_1, \mathcal{D}_3 / \mathcal{D}_2$ and $\mathcal{D}_1, \mathcal{D}_2 / \mathcal{D}_3$. For each division or batch, a gradient and a meta-gradient are back-propagated on meta-train and meta-test, respectively. By accumulating all the gradients and meta-gradients in the meta-batch, the model is finally optimized to perform well on both meta-train and meta-test domains.
Fig.~\ref{fig_metaopt} illustrates how the gradients and meta-gradients flow on the computation graph.

\begin{algorithm}
    \caption{MFR for generalized face recognition problem.}
    \label{alg_mml}
    \SetAlgoLined
    \SetKwInput{KwData}{Input}
    \SetKwInput{KwResult}{Init}
     \KwData{Source (training) domains $\mathcal{D}_S = \{\mathcal{D}_1, \mathcal{D}_2, \cdots, \mathcal{D}_N\}$.}
     \KwResult{A pre-trained model $f(\theta)$ parametrized by $\theta$, hyperparameters $\alpha$, $\beta$, $\gamma$ and batch-size of $B$.}

     \For{ite in max\_iterations}{
        Init the gradient $g_{\theta}$ as $\mathbf{0}$; \\
        Init $n$ as the number of source domains $\mathcal{D}_S$; \\
		\tcp{For a meta-batch}
        \For{each $\mathcal{D}_{mte}$ in $\mathcal{D}_S$}{
	        \tcp{For a batch}
            Sampling remaining domains as $\mathcal{D}_{mtr}$;\\
            \textbf{Meta-train:}\\
            Sampling $B$ paired images $\mathcal{X}_S$ from $B$ identities of meta-train domains $\mathcal{D}_{mtr}$;\\
            $\mathcal{L}_S {=} \mathcal{L}_{hp} (\mathcal{X}_S; \theta) + \mathcal{L}_{cls} (\mathcal{X}_S; \theta) + \mathcal{L}_{da} (\mathcal{X}_S; \theta)$; \\
            Update model parameters by: $\theta' = \theta - \alpha \nabla_{\theta} \mathcal{L}_S (\theta)$;\\
            \textbf{Meta-test:} \\
            Sampling $B$ paired images $\mathcal{X}_T$ from $B$ identities of the meta-test domain $\mathcal{D}_{mte}$;\\
            $\mathcal{L}_T = \mathcal{L}_{hp} (\mathcal{X}_T; \theta') + \mathcal{L}_{cls} (\mathcal{X}_T; \theta')$; \\
            \textbf{Gradient aggregation:} \\
            $g_{\theta} \leftarrow g_{\theta} + \gamma \nabla_{\theta} \mathcal{L}_S + (1-\gamma) \nabla_{\theta} \mathcal{L}_T$;\\
        }
        \textbf{Meta-optimization:} \\
        Update $\theta \leftarrow \theta - \frac{\beta}{n} g_{\theta}$ by $SGD$;
     }
\end{algorithm}
\section{Experiments}
To evaluate our proposed MFR for generalized face recognition problem, we conduct several experiments on two proposed benchmarks.

\begin{table*}[!htb]
    \centering
    \resizebox{0.85\textwidth}{!} {
        \begin{tabular}{cccccccc}
        \toprule[2pt]
        \multirow{2}{*}{\textbf{Facial Variety}} & \multirow{2}{*}{\textbf{Dataset}} & \multirow{2}{*}{\textbf{\#Train IDs}} & \multirow{2}{*}{\textbf{\#Train images}} & \multicolumn{2}{c}{\textbf{\#Test IDs}} & \multicolumn{2}{c}{\textbf{\#Test images}} \\ \cline{5-8} 
        &  &  &  & \textbf{\#Gallery IDs} & \textbf{\#Probe IDs} & \textbf{\#Gallery images} & \textbf{\#Probe images} \\ 
        \midrule[1pt]
        Race & Caucasian & 1,957 & 6,757 & 1,000 & 1,000 & 1,000 & 1,000 \\ 
        Race & Asian & 1,492 & 5,784 & 1,000 & 1,000 & 1,000 & 1,000 \\ 
        Race & African & 1,995 & 6,938 & 1,000 & 1,000 & 1,000 & 1,000 \\ 
        Race & Indian & 1,984 & 6,857 & 1,000 & 1,000 & 1,000 & 1,000 \\ 
		Age & CACD-VS~\cite{chen2014cross} & - & - & 2,000 & 2,000 & 2,000 & 2,000 \\ 
        Illumination & CASIA NIR-VIS 2.0~\cite{li2013casia} & - & - & 358 & 363 & 358 & 6,208 \\ 
        Pose & MultiPIE~\cite{gross2010multi} & - & - & 337 & 337 & 1,184 & 1,181 \\ 
        Occlusion & MeGlass~\cite{guo2018face} & - & - & 1,710 & 1,710 & 3,420 & 3,420 \\ 
		Heterogeneity & Public-IvS~\cite{zhu2019large} & - & - & 1,262 & 1,262 & 1,262 & 4,241 \\ 
        \bottomrule[2pt]
        \end{tabular}
    }
    \vspace{-0.5em}
    \caption{The statistics of all involved datasets. CASIA NIR-VIS 2.0 has 10 folds and the first fold is shown. The other folds own similar statistics.}
    \label{tab:dataset_stat}
    \vspace{-1.5em}
\end{table*}
\begin{table}[!htb]
    \centering
    \resizebox{0.4\textwidth}{!} {
        \begin{tabular}{l|clcc}
            \toprule[2pt]
            \multicolumn{3}{c}{\textbf{Protocol}} & \textbf{Source Domains} & \textbf{Target Domain(s)} \\ 
            \midrule[1pt]
            \multirow{12}{*}{GFR-R} & \multicolumn{2}{c|}{\multirow{3}{*}{I}} & Caucasian & \multirow{3}{*}{Indian} \\ 
            & \multicolumn{2}{c|}{} & Asian &  \\ 
            & \multicolumn{2}{c|}{} & African &  \\ \cline{2-5} 
            & \multicolumn{2}{c|}{\multirow{3}{*}{II}} & Caucasian & \multirow{3}{*}{African} \\ 
            & \multicolumn{2}{c|}{} & Asian &  \\ 
            & \multicolumn{2}{c|}{} & Indian &  \\ \cline{2-5} 
            & \multicolumn{2}{c|}{\multirow{3}{*}{III}} & Caucasian & \multirow{3}{*}{Asian} \\ 
            & \multicolumn{2}{c|}{} & African &  \\ 
            & \multicolumn{2}{c|}{} & Indian &  \\ \cline{2-5} 
            & \multicolumn{2}{c|}{\multirow{3}{*}{IV}} & Asian & \multirow{3}{*}{Caucasian} \\ 
            & \multicolumn{2}{c|}{} & African &  \\ 
            & \multicolumn{2}{c|}{} & Indian &  \\ 
            \midrule[1pt]
            \multicolumn{3}{c|}{\multirow{4}{*}{GFR-V}} & Caucasian & CACD-VS \\ 
            \multicolumn{3}{c|}{} & Asian & CASIA NIR-VIS 2.0 \\ 
            \multicolumn{3}{c|}{} & African & MultiPIE \\ 
            \multicolumn{3}{c|}{} & Indian & MeGlass \\ 
            \multicolumn{3}{c|}{} &  & Public-IvS \\
            \bottomrule[2pt]
        \end{tabular}
    }
    \vspace{-0.5em}
\caption{The GFR-R and GFR-V benchmarks. Source domains are for training, target domains are for evaluation and are unseen during training.}
\label{tab:protocol}
\end{table}

\subsection{GFR Benchmark and Protocols}
Generalized face recognition has not attracted much attention and we do not have a common protocol for evaluation, thus we introduce two well-designed benchmarks to evaluate the generalization of a model. One benchmark is for crossing race evaluation named GFR-R and another one is crossing facial variety named GFR-V. We use variety here to emphasize that there is a large gap between source domains and target unseen domains on GFR-V.

In a real-world scenario, a large-scale base dataset like MS-Celeb~\cite{guo2016ms} is usually available for pre-training, but the model may generalize poorly on a new domain with a different distribution. 
To simulate it, we use MS-Celeb as the base dataset.
RFW~\cite{wang2018racial} is originally proposed to study the racial bias in face recognition and it labels four racial datasets (Caucasian, Asian, African, Indian) from MS-Celeb. We choose to select these four datasets as our four racial domains. Note that RFW~\cite{wang2018racial} overlaps MS-Celeb~\cite{guo2016ms}, we remove all the overlapped identities from MS-Celeb according to the identity keyword, thus building our base dataset named MS-Celeb-NR\footnote{We will release the list of MS-Celeb-NR.}, which means MS-Celeb without RFW. MS-Celeb-NR can be regarded as an independent base dataset of four racial ones.

\begin{table}[!htb]
    \centering
    \resizebox{0.45\textwidth}{!} {
        \begin{tabular}{cccccc}
        \toprule[2pt]
        \multicolumn{1}{c}{\multirow{2}{*}{\textbf{Protocol}}} & \multicolumn{1}{c}{\multirow{2}{*}{\textbf{Method}}} & \multicolumn{3}{c}{\textbf{VR (\%)}} & \multicolumn{1}{c}{\multirow{2}{*}{\textbf{Rank-1 (\%)}}} \\ \cline{3-5}
        \multicolumn{1}{l}{} & \multicolumn{1}{c}{} & \multicolumn{1}{c}{\textbf{FAR=1\%}} & \multicolumn{1}{c}{\textbf{FAR=0.1\%}} & \multicolumn{1}{c}{\textbf{FAR=0.01\%}} & \multicolumn{1}{c}{} \\
        \midrule[1pt]
        \multirow{7}{*}{\textbf{\begin{tabular}[c]{@{}c@{}}GFR-R I\\ (Indian)\end{tabular}}} & Base & 94 & 82.2 & 64.65 & 80.3 \\
        & Base-Agg & 94.1 & 80.9 & 65.3 & 81 \\
        & Base-FT rnd. & 62.5 & 39 & 21.05 & 39.3 \\
        & Base-FT imp.~\cite{qi2018low} & 87 & 69.9 & 51.2 & 69.6 \\
        & MLDG~\cite{li2018learning} & 94.2 & 83 & 66.3 & 80.5 \\
        & \textbf{MFR (Ours)} & \textbf{95.4} & \textbf{86.1} & \textbf{71.4} & \textbf{83.1} \\
        \midrule[1pt]
        \multirow{7}{*}{\textbf{\begin{tabular}[c]{@{}c@{}}GFR-R II\\ (African)\end{tabular}}} & Base & 91.6 & 74.5 & 55.4 & 73.1 \\
        & Base-Agg & 90.5 & 74.8 & 56.3 & 74 \\
        & Base-FT rnd. & 26.2 & 10.9 & 3.5 & 21 \\
        & Base-FT imp.~\cite{qi2018low} & 78.7 & 56.6 & 36.45 & 57.9 \\
        & MLDG~\cite{li2018learning} & 91.9 & 74.8 & 55.7 & 73.8 \\
        & \textbf{MFR (Ours)} & \textbf{92.3} & \textbf{79.4} & \textbf{60.8} & \textbf{75.2} \\
        \midrule[1pt]
        \multirow{7}{*}{\textbf{\begin{tabular}[c]{@{}c@{}}GFR-R III\\ (Asian)\end{tabular}}} & Base & 91.89 & 77.98 & 60.86 & 75.98 \\
        & Base-Agg & 91.49 & 78.08 & 59.41 & 76.28 \\
        & Base-FT rnd. & 40.44 & 17.32 & 7.67 & 27.53 \\
        & Base-FT imp.~\cite{qi2018low} & 80.58 & 57.56 & 39.79 & 61.86 \\
        & MLDG~\cite{li2018learning} & 92.29 & 78.28 & 60.3 & 76.68 \\
        & \textbf{MFR (Ours)} & \textbf{93.49} & \textbf{80.7} & \textbf{62.56} & \textbf{78.68} \\
        \midrule[1pt]
        \multirow{7}{*}{\textbf{\begin{tabular}[c]{@{}c@{}}GFR-R IV\\ (Caucasian)\end{tabular}}} & Base & 96.6 & 89.6 & 78.6 & 86.6 \\
        & Base-Agg & 97 & 88.1 & 79.1 & 86.8 \\
        & Base-FT rnd. & 61.1 & 36.2 & 18.9 & 36.7 \\
        & Base-FT imp.~\cite{qi2018low} & 91.5 & 78.2 & 63.4 & 76.8 \\
        & MLDG~\cite{li2018learning} & 96.8 & 89.6 & 79.15 & 86.3 \\
        & \textbf{MFR (Ours)} & \textbf{98.2} & \textbf{92.9} & \textbf{81.1} & \textbf{88.9} \\
        \bottomrule[2pt]
        \end{tabular}
    }
    \vspace{-0.5em}
    \caption{Comparative results of the GFR-R benchmark. rnd. means random initializing the classification weight template, imp. is weight-imprinted.}
    \label{tab:gfr_r}
\end{table}

\textbf{GFR-R.} 
Each race has about 2K$\sim$3K identities. We randomly choose 1K identities for testing and the remaining 1K$\sim$2K identities for training. The dataset details are shown in Table~\ref{tab:dataset_stat}. In our experiment setting, each race is regarded as one domain. We randomly select three domains in four as source domains and and the rest one as the testing domain, which is not accessible in training. Therefore, we build four sub-protocols for GFR-R, shown in Table~\ref{tab:protocol}.

\textbf{GFR-V.} The GFR-V benchmark is for crossing facial variety evaluation, which is a harder setting and can better reflect the generalization ability of a model. As is shown in Table~\ref{tab:protocol}, four racial datasets (Caucasian, Asian, African, Indian) are treated as source domains, and five datasets are as target domains. Specifically, the target datasets include CACD-VS~\cite{chen2014cross}, CASIA NIR-VIS 2.0~\cite{li2013casia}, MultiPIE~\cite{gross2010multi}, MeGlass~\cite{guo2018face}, Public-IvS~\cite{zhu2019large}. For CASIA NIR-VIS 2.0, we follow the standard protocol in View 2 evaluation~\cite{li2013casia} and we report the average value of 10 folds. For MeGlass and Public-IvS, we follow the standard testing protocols~\cite{zhu2019large,guo2018face}. For CACD-VS, in addition to the standard protocol~\cite{chen2014cross}, we use the provided 2,000 positive cross-age image pairs and split them into gallery and probe for our ROC/Rank-1 evaluation.  For Multi-PIE, we select 337 identities and each identity contains about 3$\sim$4 frontal gallery images and 3$\sim$4 probe images with the 45$^\circ$ view.

\textbf{Benchmark Protocols.} For each image, the features from both the original image and the flipped one are extracted then concatenated as the final representation. The score is measured by the cosine distance of two representations.
For performance evaluation, we use the receiver operating characteristic (ROC) curve and Rank-1 accuracy. For ROC, we report the verification rate (VR) at low false acceptance rate (FAR) like 1\%, 0.1\% and 0.01\%. For Rank-1 evaluation, each probe image is matched to all gallery images, if the top-1 result is within the same identity, it is correct.

\subsection{Implementation Details}
Our experiments are based on PyTorch~\cite{paszke2017automatic}. The random seed is set to a fixed value 2019 in comparative experiments for fair comparisons.
We use a 28-layer ResNet as our backbone, but with a channel-number multiplier of 0.5.
Our backbone has only 128.7M FLOPs and 4.64M parameters, which is relatively light-weighted.
The dimension of the output embedding is 256.
The model is pre-trained on MS-Celeb-NR with CosFace~\cite{wang2018cosface}.
During training, all faces are cropped and resized to 120$\times$120. The inputs are then normalized by subtracting 127.5 and being divided by 128.
The meta-train step-size $\alpha$, the meta optimization step-size $\beta$, the weight $\gamma$ balancing meta-train and meta-test loss are initialized to 0.0004, 0.0004, 0.5, respectively. Batch-size $B$ is set to 128 and the scaling factor $s$ of both the soft-classification loss and domain alignment loss are set to 64.
The step-size $\alpha$ and $\beta$ are decayed with every 1K steps and the decay rate is 0.5. The positive threshold $\tau_p$ and negative threshold $\tau_n$ are initialized to 0.3, 0.04 and are updated as $\tau_p = 0.3 + 0.1 n$ and $\tau_n = 0.04 / 0.5 ^ n$, where n is the decayed number.
For meta-optimization, we use SGD to optimize the network with the weight decay of 0.0005 and momentum of 0.9.

\subsection{GFR-R Comparisons}

\textbf{Settings.}
We compare our model with several baselines, including the base model and several domain aggregation baselines.
To further compare our method with other domain generalization methods, we adapt MLDG~\cite{li2018learning} to an open-set setting, so that it can be applied in our protocols. 
The results are shown in Table~\ref{tab:gfr_r}. For four protocols in GFR-R, we report the VRs at low FAR 1\%, 0.1\%, 0.01\%, and the Rank-1 accuracy. Specifically, our comparisons include: (i) \textit{Base}: the model pre-trained only on MS-Celeb-NR using CosFace~\cite{wang2018cosface}. Note that MS-Celeb-NR has no overlapped identities with four racial datasets (Caucasian, Asian, African and Indian) and can be considered as an independent dataset. (ii) \textit{Base-Agg}: the model trained on MS-Celeb-NR and the aggregation of source domains using CosFace~\cite{wang2018cosface}. Take GFR-R-I as an example, \textit{Base-Agg} is trained on MS-Celeb-NR and three source domains Caucasian, Asian, African jointly. This is for the fair comparison with our MFR, where the same training datasets are involved. (iii) \textit{Base-FT rnd.}: the base model further fine-tuned on the aggregation of source domains. The classification template of the last FC-layer is randomly initialized. (iv) \textit{Base-FT imp.}: the base model further fine-tuned on the aggregation of source domains, but the classification template is initialized as the mean of embeddings of the corresponding identities. It is refined from weight-imprinted~\cite{qi2018low}. (v) \textit{MLDG}: MLDG~\cite{li2018learning} adapted for generalized face recognition problem.

\textbf{Results.} From the results in Table~\ref{tab:gfr_r}, the following observations can be made: (i) Overall, our method achieves the best result on four GFR-R protocols among all compared settings and methods. (ii) The base model pre-trained on MS-Celeb-NR is strong, but not generalizes  well for target domains, especially for Indian, African, Asian. The reason may be that MS-Celeb-NR is occupied by Caucasian people. (iii) Jointly training on MS-Celeb-NR and source domains performs slightly better than the base model, but is still not comparable to our MFR method. (iv) The performance of \textit{Base-FT rnd.} declines dramatically and we attribute it to over-fitting on source domains. Weight-imprinted (\textit{Base-FT imp.}) can reduce such over-fitting to some degree, but its performance is still lower than the base model. (v) MLDG~\cite{li2018learning}, which is originally designed for closed-set and category-level recognition problems, fail to compete with our method on the open-set generalized face recognition problem.

\begin{table}[!htb]
    \centering
    \resizebox{0.475\textwidth}{!} {
        \begin{tabular}{cccccc}
        \toprule[2pt] 
        \multirow{2}{*}{\textbf{\begin{tabular}[c]{@{}c@{}}GFR-V\\ (CACD-VS)\end{tabular}}} & \multicolumn{2}{c}{\textbf{VR (\%)}} & \multirow{2}{*}{\textbf{Rank-1 (\%)}} & \multirow{2}{*}{\textbf{Acc.}} & \multirow{2}{*}{\textbf{AUC.}} \\ \cline{2-3}
 & \textbf{FAR=0.01\%} & \textbf{FAR=0.001\%} &  &  &  \\ \midrule[1pt] 
        Base & 96.55 & 92.55 & 96.85 & 99.35 & 99.53 \\ 
        Base-Agg & 96.75 & 92.98 & 97.15 & 99.42 & 99.6 \\ 
        MLDG~\cite{li2018learning} & 96.75 & 92.9 & 97.25 & 99.45 & 99.57 \\
        LF-CNNs~\cite{wen2016latent} & - & - & - & 98.5 & 99.3 \\
        Human, Voting~\cite{chen2015face}  & - & - & - & 94.2 & 99 \\
        OE-CNNs~\cite{wang2018orthogonal} & - & - & - & 99.2 & 99.5 \\ 
        AIM+CAFR~\cite{zhao2019look} & - & - & - & 99.76 & - \\
        \textbf{MFR (Ours)} & \textbf{97.25} & \textbf{94.05} & \textbf{97.8} & \textbf{99.78} & \textbf{99.81} \\ 
        \bottomrule[2pt]
        \end{tabular}
    }
    \vspace{-0.5em}
    \caption{Comparative results on CACD-VS.}
    \vspace{-0.5em}
    \label{tab:cacdvs}
\end{table}

\begin{table}[]
    \centering
    \resizebox{0.475\textwidth}{!} {
        \begin{tabular}{ccccc}
        \toprule[2pt] 
        \multirow{2}{*}{\textbf{\begin{tabular}[c]{@{}c@{}}GFR-V\\ CASIA NIR-VIS 2.0\end{tabular}}} & \multicolumn{3}{c}{\textbf{VR (\%)}} & \multirow{2}{*}{\textbf{Rank-1 (\%)}} \\ \cline{2-4}
        & \textbf{FAR=1\%} & \textbf{FAR=0.1\%} & \textbf{FAR=0.01\%} &  \\  \midrule[1pt] 
        Base & 97.8 & 89.89 & 69.27 & 93.18 \\ 
        Base-Agg & 98.31 & 90.47 & \textbf{71.35} & 94.29 \\ 
        MLDG~\cite{li2018learning} & 98.28 & 90.44 & 69.32 & 93.56 \\
        IDR~\cite{he2017learning} & 98.9 & 95.7 & - & \textbf{97.3} \\
        WCNN~\cite{he2018wasserstein} & \textbf{99.4} & \textbf{97.6} & - & \textbf{98.4} \\
        \textbf{MFR (Ours)} & \textbf{99.32} & \textbf{95.97} & \textbf{81.92} & 96.92 \\  
        \bottomrule[2pt]
        \end{tabular}
    }
    \vspace{-0.5em}
    \caption{Comparative results on CASIA NIR-VIS 2.0. The highest two results are highlighted.}
    \vspace{-0.5em}
    \label{tab:nirvis2}
\end{table}

\begin{table}[]
    \centering
    \resizebox{0.475\textwidth}{!} {
        \begin{tabular}{ccccc}
        \toprule[2pt]
        \multirow{2}{*}{\textbf{\begin{tabular}[c]{@{}c@{}}GFR-V\\ (Multi-PIE)\end{tabular}}} & \multicolumn{3}{c}{\textbf{VR (\%)}} & \multirow{2}{*}{\textbf{Rank-1 (\%)}} \\ \cline{2-4}
        & \textbf{FAR=0.1\%} & \textbf{FAR=0.01\%} & \textbf{FAR=0.001\%} &  \\ \midrule[1pt]
        Base & 99.92 & 98.83 & 61.54 & 99.75 \\
        Base-Agg & 99.92 & 98.96 & 68.49 & 99.82 \\
        MLDG~\cite{li2018learning} & 99.84 & 98.87 & 62.95 & 99.83 \\
        \textbf{MFR (Ours)} & \textbf{100} & \textbf{99.96} & \textbf{74.54} & \textbf{99.92} \\
        \bottomrule[2pt]
        \end{tabular}
    }
    \vspace{-0.5em}
    \caption{Comparative results on MultiPIE.}
    \vspace{-0.5em}
    \label{tab:multipie}
\end{table}

\begin{table}[]
    \centering
    \resizebox{0.475\textwidth}{!} {
        \begin{tabular}{ccccc}
            \toprule[2pt]
        \multirow{2}{*}{\textbf{\begin{tabular}[c]{@{}c@{}}GFR-V\\ (MeGlass)\end{tabular}}} & \multicolumn{3}{c}{\textbf{VR (\%)}} & \multirow{2}{*}{\textbf{Rank-1 (\%)}} \\ \cline{2-4}
        & \textbf{FAR=0.01\%} & \textbf{FAR=0.001\%} & \textbf{FAR=0.0001\%} &  \\ \midrule[1pt]
        Base & 85.92 & 71.96 & 53.5 & 97.6 \\
        Base-Agg & 86.77 & 73.5 & 54.96 & 97.69 \\
        MLDG~\cite{li2018learning} & 85.54 & 69.23 & 49.32 & \textbf{97.81} \\
        Face Syn.~\cite{guo2018face} & \textbf{90.14} & \textbf{80.32} & \textbf{66.92} & 96.73 \\
        \textbf{MFR (Ours)} & \textbf{90.79} & \textbf{80.86} & \textbf{66.15} & \textbf{98.57} \\
        \bottomrule[2pt]
        \end{tabular}
    }
    \vspace{-0.5em}
    \caption{Comparative results on MeGlass. The highest two results are highlighted.}
    \vspace{-0.5em}
    \label{tab:meglass}
\end{table}

\begin{table}[]
    \centering
    \resizebox{0.45\textwidth}{!} {
        \begin{tabular}{ccccc}
            \toprule[2pt]
        \multirow{2}{*}{\textbf{\begin{tabular}[c]{@{}c@{}}GFR-V\\ (Public-IvS)\end{tabular}}} & \multicolumn{3}{c}{\textbf{VR (\%)}} & \multirow{2}{*}{\textbf{Rank-1 (\%)}} \\ \cline{2-4}
        & \textbf{FAR=0.1\%} & \textbf{FAR=0.01\%} & \textbf{FAR=0.001\%} &  \\ \midrule[1pt]
        Base & 94.38 & 86.71 & 74.83 & 92.74 \\
        Base-Agg & 94.24 & 87.1 & 74.5 & 92.85 \\
        MLDG~\cite{li2018learning} & 94.96 & 87.35 & 75.54 & \textbf{93.3} \\
        Contrastive~\cite{sun2014deep_iv} & 96.52 & 91.71 & 84.54 & - \\
        LBL~\cite{zhu2019large} & \textbf{98.83} & \textbf{97.21} & \textbf{93.62} & - \\
        \textbf{MFR (Ours)} & \textbf{96.66} & \textbf{92.96} & \textbf{85.28} & \textbf{95.82} \\
        \bottomrule[2pt]
        \end{tabular}
    }
    \vspace{-0.5em}
    \caption{Comparative results on Public-IvS. The highest two results are highlighted.}
    \vspace{-0.5em}
    \label{tab:publicivs}
\end{table}


\begin{table}[!htb]
    \centering
    \resizebox{0.425\textwidth}{!} {
        \begin{tabular}{ccccc}
        \toprule[1pt]
        \textbf{Method} & Base & Base-Agg & MLDG~\cite{li2018learning} & \textbf{MFR (Ours)} \\ 
        \textbf{LFW}    &  99.57    &    99.60      &  99.43    &  \textbf{99.77}         \\
        \bottomrule[1pt]
        \end{tabular}
    }
    \vspace{-0.5em}
    \caption{Comparative results on LFW.}
    \label{tab:lfw}
    \vspace{-1.5em}
\end{table}


\subsection{GFR-V Comparisons}
The GFR-V benchmark is for crossing facial variety evaluation, which can better reflect model generalization.

\textbf{Settings.} We compare our model with two strong baselines \textit{Base}, \textit{Base-Agg}, an adapted MLDG~\cite{li2018learning} and other competitors if existed. Since the standard protocols differ among five target domains, we show them separately in Table~\ref{tab:cacdvs},~\ref{tab:nirvis2},~\ref{tab:multipie},~\ref{tab:meglass},~\ref{tab:publicivs}.

\textbf{CACD-VS.} CACD-VS~\cite{chen2014cross} is for cross-age evaluation, where each pair of images contain a young face and an old one. We report ROC/Rank-1 as well as the standard protocol provided. Other competitors are only evaluated on the standard protocol. The results in Table~\ref{tab:cacdvs} show that our MFR not only beats the baselines but also the competitors, which use cross-age datasets for training.

\textbf{CASIA NIR-VIS 2.0.} In CASIA NIR-VIS 2.0~\cite{li2013casia}, gallery images are collected under visible lighting, while the probe one is under near infrared lighting, thus the modality gap is huge. Table~\ref{tab:nirvis2} shows: (i) we achieve great performance improvements from 89.89\% (69.27\%) of \textit{Base} to 95.97\% (81.92\%) when FAR=0.1\% (0.01\%). (ii) even with such a huge modality gap, our performance is comparable to several CNN-based methods~\cite{he2017learning,he2018wasserstein}, which use MS-Celeb for pre-training and the target domain NIR-VIS dataset for fine-tuning. In comparison, our model has not seen any NIR samples during training.

\textbf{Multi-PIE.} We compare our model with two baselines and MLDG for cross-pose evaluation using Multi-PIE. Table~\ref{tab:multipie} validates the improvements of our MFR over baselines and MLDG.

\textbf{MeGlass.} MeGlass~\cite{guo2018face} focuses on the effect of eyeglass occlusion for face recognition. We select the hardest IV protocol for evaluation. As shown in Table~\ref{tab:meglass}, our method promotes the performance from 71.96\% (53.5\%) on \textit{Base} to 80.86\% (66.15\%) when at a low FAR 0.001\%, which is even slightly better than ~\cite{guo2018face}, which synthesizes wearing-eyeglass image for the whole MS-Celeb for training.

\textbf{Public-IvS.} Public-IvS~\cite{zhu2019large} is a testbed for ID vs. Spot (IvS) verification. Compared to \textit{Base} and \textit{Base-Agg}, our method greatly improves the generalization performance. The other two competitors are all pre-trained on MS-Celeb and fine-tuned on CASIA-IvS, which has more than 2 million identities and each identity has one ID and Spot face. Even so, our method still performs slightly better than Contrastive~\cite{sun2014deep_iv}.

\textbf{LFW.} We perform an extensive evaluation on LFW~\cite{huang2008labeled}, shown in Table.~\ref{tab:lfw}. The results demonstrate that our method also generalizes better than baselines on a similar target domain.

The above results show that our method achieves great improvement than baselines, and the performance is competitive to the best supervised / non-generalization methods. For a real-world face recognition application, our method is the first choice because it generalizes well on all target domains with competitive performances.

\begin{table}[!htb]
    \centering
    \vspace{-0.5em}
    \resizebox{0.425\textwidth}{!} {
        \begin{tabular}{cccccc}
        \toprule[2pt]
        \multicolumn{1}{c}{\multirow{2}{*}{\textbf{Protocol}}} & \multicolumn{1}{c}{\multirow{2}{*}{\textbf{Method}}} & \multicolumn{3}{c}{\textbf{VR (\%)}} & \multicolumn{1}{c}{\multirow{2}{*}{\textbf{Rank-1 (\%)}}} \\ \cline{3-5}
        \multicolumn{1}{l}{} & \multicolumn{1}{c}{} & \multicolumn{1}{c}{\textbf{FAR=1\%}} & \multicolumn{1}{c}{\textbf{FAR=0.1\%}} & \multicolumn{1}{c}{\textbf{FAR=0.01\%}} & \multicolumn{1}{c}{} \\
        \midrule[1pt]
        \multirow{5}{*}{\textbf{\begin{tabular}[c]{@{}c@{}}GFR-R I\\ (Indian)\end{tabular}}} 
        & w/o hp. & 95.1 & 84.5 & 69.2 & 82.2 \\
        & w/o cls. & 95.3 & 84.3 & 69 & 82.3 \\
        & w/o da. & 95.2 & 84.9 & 70.8 & 82.7 \\
        & w/o meta ($\alpha=0$) & 94.8 & 84.3 & 68.35 & 81.1 \\
        & first order & 95.3 & 85.7 & 70.9 & 82.6 \\
        & \textbf{Ours-full} & \textbf{95.4} & \textbf{86.1} & \textbf{71.4} & \textbf{83.1} \\
        \midrule[1pt]
        \multirow{5}{*}{\textbf{\begin{tabular}[c]{@{}c@{}}GFR-R II\\ (African)\end{tabular}}} 
        & w/o hp. & 92 & 77.9 & 59.2 & 74.6 \\
        & w/o cls. & 92.1 & 78.9 & 59.3 & 74.6 \\
        & w/o da. & 92.1 & 78.6 & 59.4 & 74.8 \\
        & w/o meta ($\alpha=0$) & 91.9 & 77.6 & 57.75 & 74.8 \\
        & first order & 92 & 78.05 & 59.5 & \textbf{75.2} \\
        & \textbf{Ours-full} & \textbf{92.3} & \textbf{79.4} & \textbf{60.8} & \textbf{75.2} \\
        \midrule[1pt]
        \multirow{5}{*}{\textbf{\begin{tabular}[c]{@{}c@{}}GFR-R III\\ (Asian)\end{tabular}}} 
        & w/o hp. & 93.39 & 79.9 & 61.56 & 77.78 \\
        & w/o cls. & 93.29 & 80.4 & 62.1 & 78.08 \\
        & w/o da. & 93.49 & 79.68 & 61.76 & 77.78 \\
        & w/o meta ($\alpha=0$) & 92.89 & 79.2 & 60.7 & 77.28 \\
        & first order & \textbf{93.49} & 79.9 & 61.7 & 77.88 \\
        & \textbf{Ours-full} & \textbf{93.49} & \textbf{80.7} & \textbf{62.56} & \textbf{78.68} \\
        \midrule[1pt]
        \multirow{5}{*}{\textbf{\begin{tabular}[c]{@{}c@{}}GFR-R IV\\ (Caucasian)\end{tabular}}} 
        & w/o hp. & 98.2 & 91.3 & 80.4 & 87.8 \\
        & w/o cls. & 98.3 & 92.6 & 80.4 & 88.4 \\
        & w/o da. & \textbf{98.4} & 92.4 & 80.5 & 87.5 \\
        & w/o meta ($\alpha=0$) & 97.3 & 91.1 & 79.6 & 87.3 \\
        & first order & 97.9 & 91.8 & 80.1 & 87.7 \\
        & \textbf{Ours-full} & 98.2 & \textbf{92.9} & \textbf{81.1} & \textbf{88.9} \\
        \bottomrule[2pt]
        \end{tabular}
    }
    \vspace{-0.5em}
    \caption{Ablative results of the GFR-R benchmark. hp. is the hard-pair attention loss, cls. is the soft-classification loss and da. is the domain alignment loss on meta-train domains.}
    \vspace{-1.5em}
    \label{tab:gfr_r_ablative}
\end{table}

\subsection{Ablation Study and Analysis}
\textbf{Contribution of Different Components.} 
To evaluate the contributions of different components, we compare our full MFR with four degraded versions.
The first three components are the hard-pair attention loss, soft-classification loss and domain alignment loss, which are designed for learning domain-invariant and discriminative representations. The fourth component is the meta-gradient. If $\alpha$ is set to 0 in Eqn.~\ref{eq_finalobj}, the objective is degraded to the sum of meta-train and meta-test and there is no meta-gradient computation.
Table~\ref{tab:gfr_r_ablative} shows that each component contributes to the performance. Among three components, the meta-gradient is the most important one. For example, in GFR-R I, the performance drops from 71.4\% to 68.35\% when FAR=0.01\% without the meta-gradient.

\begin{figure}[!htb]
    \centering
    \includegraphics[width=0.35\textwidth]{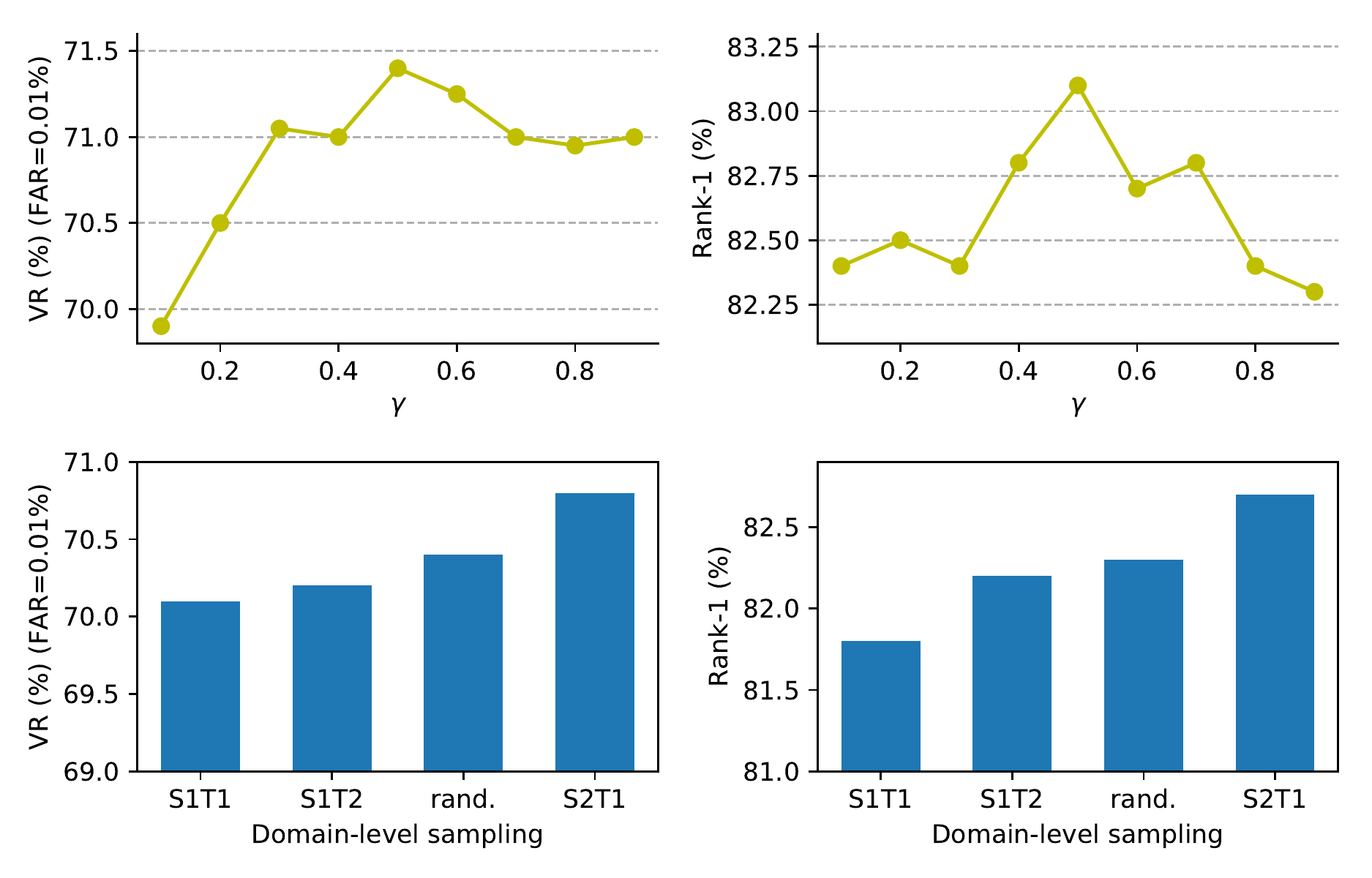}
    \vspace{-1em}
    \caption{Ablation results on GFR-R I (Indian) protocol, with different $\gamma$ and domain-level sampling strategies.} 
    \vspace{-1.5em}
    \label{fig_ablation_gamma}
\end{figure}

\textbf{First Order Approximation.}
The meta-gradient needs high order derivatives and is computationally expensive. Therefore, we compare it with the first order approximation.
To achieve the first order approximation, we only need to change $(1-\gamma) \nabla_{\theta} \mathcal{L}_T$ in Algorithm~\ref{alg_mml} to $(1-\gamma) \nabla_{\theta'} \mathcal{L}_T$ in the gradient aggregation step.
From Table~\ref{tab:gfr_r_ablative}, we can see that the performance of the first order approximation is close to high order. Considering that the first order approximation takes only about 82\% GPU memory and 63\% time (in our setting) of the high order, the first order approximation is a practical substitute for the high order implementation.

\textbf{Impact of $\gamma$.} In Eqn.~\ref{eq_finalobj}, $\gamma$ is a hyperparameter weighting the meta-train and meta-test losses. The ablative results are shown in Fig.~\ref{fig_ablation_gamma}. A proper value 0.5 gives the best result, which indicates the meta-train and meta-test domains should be equally learned.

\textbf{Domains-level Sampling.} Since domain alignment loss cannot be applied when there is only one domain in meta-train, we remove it for fair comparisons. For each batch, S$m$T$n$ ( $m,n \in \{ (1,1), (1,2),  (2,1) \}$ ) means sampling $m$ domain as meta-train and another $n$ as meta-test. rand. means randomly choosing $m$ domains as meta-train ($m$ is a random number) and remaining one as meta-test. Fig.~\ref{fig_ablation_gamma} shows that the setting $m=2$ and $n=1$ performs best.

\section{Conclusion}
In this paper, we highlight generalized face recognition problem and propose a Meta Face Recognition (MFR) method to address it. Once trained on a set of source domains, the model can be directly deployed on target domains without any model update.
Extensive experiments on two newly defined generalized face recognition benchmarks validate the effectiveness of our proposed MFR.
We believe generalized face recognition problem is of great importance for practical applications, and our work is an important avenue for future works.
\section*{Acknowledgments}
This work has been partially supported by the Chinese National Natural Science Foundation Projects \#61876178, \#61806196, \#61976229, \#61872367 and Science and Technology Development Fund of Macau (No.~0008/2018/A1, 0025/2019/A1, 0019/2018/ASC, 0010/2019/AFJ, 0025/2019/AKP).

{\small
\bibliographystyle{unsrt}
\bibliography{egbib}
}

\end{document}